\documentclass[10pt,twocolumn,letterpaper]{article}

\usepackage{cvpr}
\usepackage{times}
\usepackage{epsfig}
\usepackage{graphicx}
\usepackage{amsmath}
\usepackage{amssymb}
\usepackage{booktabs}
\usepackage {multirow}
\usepackage{fixltx2e}

\newcommand{\vect}[1]{\mathbf{#1}}

\def\eg{\textit{e.g.}~}
\def\ie{\textit{i.e.}~}

\def\etal{\textit{et al.}~}

\usepackage{tabularx}

\usepackage{etoolbox}
\makeatletter
\patchcmd{\maketitle}
 {\def\@makefnmark}
 {\def\@makefnmark{}\def\useless@macro}
 {}{}
\makeatother

\newcommand{\tabincell}[2]{\begin{tabular}{@{}#1@{}}#2\end{tabular}}

\usepackage[pagebackref=true,breaklinks=true,letterpaper=true,colorlinks,bookmarks=false]{hyperref}

\cvprfinalcopy 

\def\cvprPaperID{3491} 

\ifcvprfinal\fi
\begin{document}

\title{Structured Attention Guided 
Convolutional Neural Fields \\ for 
Monocular Depth Estimation\vspace{-15pt}}

\author{Dan Xu$^{1}\thanks{corresponding author.}$, \ Wei Wang$^1$, \ Hao Tang$^1$, \ Hong Liu$^{2*}$, \ Nicu Sebe$^1$, \ Elisa Ricci$^{1,3*}$\\
$^1$Multimedia and Human Understanding Group, University of Trento, \\ $^2$Key Laboratory of Machine Perception, Shenzhen Graduate School, Peking University, \\
$^3$ Technologies of Vision Group, Fondazione Bruno Kessler
\\
{\tt\small \{dan.xu, wei.wang, hao.tang, niculae.sebe, e.ricci\}@unitn.it \ hongliu@pku.edu.cn} 
\vspace{-15pt}
}
\maketitle
\begin{abstract}
Recent works have shown the benefit of integrating Conditional Random Fields (CRFs) models into deep architectures
for improving pixel-level prediction tasks. Following this line of research, in this paper
we introduce a novel approach for monocular depth estimation. Similarly to previous works, our method employs
a continuous CRF to fuse multi-scale information derived from different layers of a front-end Convolutional Neural Network (CNN).
Differently from past works, our approach benefits from a structured attention model which automatically regulates  
the amount of
information transferred between corresponding features at different scales. Importantly, the proposed attention model is
seamlessly integrated into the CRF, allowing end-to-end training of the entire architecture.
Our extensive experimental evaluation demonstrates the effectiveness of the proposed method
which is competitive with previous methods on the KITTI benchmark and outperforms the state of the art on the NYU Depth V2 dataset.
\end{abstract}

\section{Introduction}
The problem of recovering depth information from images has been widely studied in computer
vision. Traditional approaches operate by considering multiple observations
of the scene of interest, \eg derived from two or more cameras or corresponding
to different lighting conditions. More recently, 
the research community has attempted to relax the multi-view assumption by
addressing the task of monocular depth estimation
as a supervised learning problem. Specifically,
given a large training set of pairs of images and associated depth maps,
depth prediction is casted as a pixel-level regression problem, \ie a
model is learned to directly predict
the depth value corresponding to each pixel of an RGB image.

\begin{figure}[!t]
\centering
\includegraphics[width=3.2in]{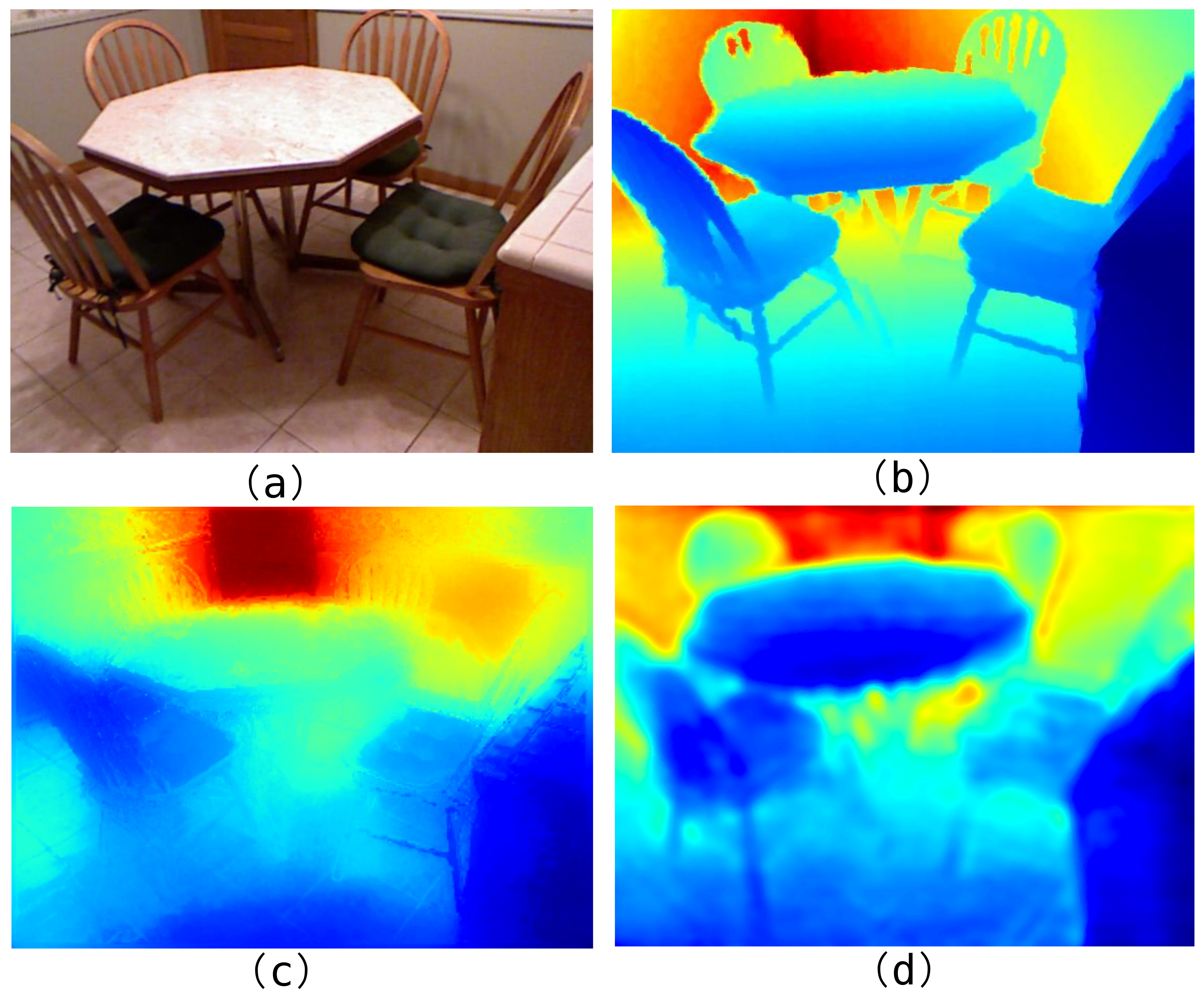} 
\caption{Monocular depth prediction from different CRF-based models: Xu~\etal~\cite{xu2017multi} (c) and ours (d). (a) and (b) are the input RGB 
image and the corresponding ground truth depth map.} 
\label{motivation}
\vspace{-15pt}
\end{figure}

In the last few years several approaches have 
been proposed for addressing this task and
remarkable performance has been achieved thanks to
deep learning models \cite{eigen2015predicting,eigen2014depth,liu2015deep,xu2017multi,kuznietsov2017semi}. 
Recently, various Convolutional Neural Network (CNN) architectures have been proposed, tackling different sub-problems
such as how to jointly estimate depth maps and semantic labels~\cite{xu2018PAD-Net}, how to build models robust to noise or how to combine multi-scale features~\cite{hariharan2015hypercolumns}. Focusing on the latter issue, 
recent works have shown that CRFs can be integrated into deep architectures \cite{liu2015deep,wang2015towards} and can be exploited to optimally fuse the multi-scale information derived from inner layers of a CNN \cite{xu2017multi}.

Inspired by these works, in this paper we also propose to exploit the flexibility 
of graphical models for multi-scale monocular depth estimation. However, we significantly depart from
previous methods and we argue that more accurate estimates can be obtained operating not only
at the prediction level but exploiting directly the internal CNN feature representations. To this aim, we design a 
novel CRF model which automatically learns robust multi-scale features by
integrating an attention mechanism. 
Our attention model allows to automatically regulate how much information should flow
between related features at different scales.


Attention models have been successfully adopted in computer vision and they have shown to be especially useful
for improving the performance of CNNs in pixel-level prediction tasks, such as semantic segmentation \cite{chen2015attention,hong2015TransferNet}.
In this work we demonstrate that attention models are also extremely beneficial in the context of monocular depth prediction.
We also show that the attention variables can be jointly estimated with multi-scale feature representations during CRF inference and that, by
employing a structured attention model \cite{kim2017structured} (\ie by imposing similarity constraints
between attention variables for related pixels and scales), we can further boost performance.
Through extensive experimental evaluation
we demonstrate that our method produces more accurate depth maps than
traditional approaches based on CRFs \cite{liu2015deep,wang2015towards} and multi-scale CRFs \cite{xu2017multi} (Fig.\ref{motivation}). 
Moreover, by performing experiments on the publicly available NYU Depth V2 \cite{silberman2012indoor} and on the 
KITTI \cite{Geiger2013IJRR} datasets, we show that our approach outperforms most state of the art methods. 

\vspace{-5mm}

\paragraph{Contributions.} In summary, we make the following contributions: (i) We
propose a novel deep learning model for calculating depth maps from still images which seamlessly integrates
a front-end CNN and a multi-scale CRF. Importantly, our model
can be trained end-to-end. Differently from previous works \cite{xu2017multi,liu2015deep,wang2015towards} {our framework does not consider as input only 
prediction maps but operates directly at feature-level. Furthermore, by adopting appropriate unary and pairwise potentials, 
our framework allows a much faster inference}. 
(ii) Our approach benefits from a novel attention mechanism which
allows to robustly fuse features derived from multiple scales as well as to integrate structured information.
(iii) Our method demonstrates state-of-the-art performance on the NYU Depth V2 \cite{silberman2012indoor} dataset and is among the top performers
on the more challenging outdoor scenes of the KITTI benchmark \cite{Geiger2013IJRR}. The code is made publicly available\footnote{https://github.com/danxuhk/StructuredAttentionDepthEstimation}.

\section{Related work}
\paragraph{Monocular Depth Estimation.} The problem of monocular depth estimation has
attracted considerable attention in last decade. While earlier approaches are mostly based on 
hand-crafted features \cite{hoiem2005automatic,karsch2014depth,ladicky2014pulling,saxena20083},
more recent works adopt deep architectures \cite{eigen2015predicting,liu2015deep,wang2015towards,roymonocular,laina2016deeper,xu2017multi,godard2016unsupervised}.
In \cite{eigen2014depth} a model based on two CNNs 
is proposed: a first network is used for estimating depth at a coarse scale,
while the second one is adopted to refine predictions. In \cite{laina2016deeper} a residual network
integrating a novel reverse Huber loss is presented. In \cite{cao2016} a deep residual network
is also employed but the problem of depth estimation from still images is translated from a regression to a classification task.
Recent works have also shown the benefit of adopting multi-task learning strategies, \eg for jointly predicting depth and performing semantic segmentation, ego-motion estimation or surface normal computation \cite{eigen2015predicting,zhou2017unsupervised,wang2015towards}.
Some recent papers have proposed unsupervised or weakly supervised methods for reconstructing depth maps \cite{godard2016unsupervised,kuznietsov2017semi}.
Other works have exploited the flexibility of graphical models within deep learning architectures
for estimating depth maps. For instance, in \cite{wang2015towards} a Hierarchical CRF is adopted to refine 
depth predictions obtained by a CNN. In \cite{liu2015deep} a continuous CRF is proposed for generating depth maps
from CNN features computed on superpixels. The most similar work to ours is \cite{xu2017multi}, where a CRF is adopted to combine multi-scale information derived from multiple inner layer of a CNN. Our approach develops from a similar intuition but further integrates an attention model which significantly improves the accuracy of the estimates. To our knowledge this is the first paper exploiting attention mechanisms in the context of monocular depth estimation.

\vspace{-5mm}

\paragraph{Fusing Multi-scale Information in CNNs.} Many recent works have shown the benefit of combining multi-scale information
for pixel-level prediction tasks such as semantic segmentation \cite{chen2014semantic}, 
depth estimation \cite{xu2017multi} or contour detection \cite{xie2015holistically}. For instance,
dilated convolutions are employed in \cite{chen2014semantic}. Multi-stream architectures with inputs at different
resolutions are considered in \cite{buyssens2012multiscale}, while \cite{long2015fully} proposed skip-connections
to fuse feature maps derived from different layers. In \cite{xie2015holistically} deep supervision
is exploited for fusing information from multiple inner layers. 
CRFs have been considered for integrating multi-scale information
in \cite{xu2017multi}. In \cite{chen2015attention, xu2017learning} an attention model is employed for combining multi-scale features in the context of semantic segmentation and object contour detection. The approach we present in this paper is radically different, 
as we employ a structured attention model which is jointly learned within a CRF-CNN framework.

%


\section{Estimating Depth Maps with Structured 
Attention Guided Conditional Neural Fields}

\begin{figure*}[t]
\centering
\includegraphics[width=6in]{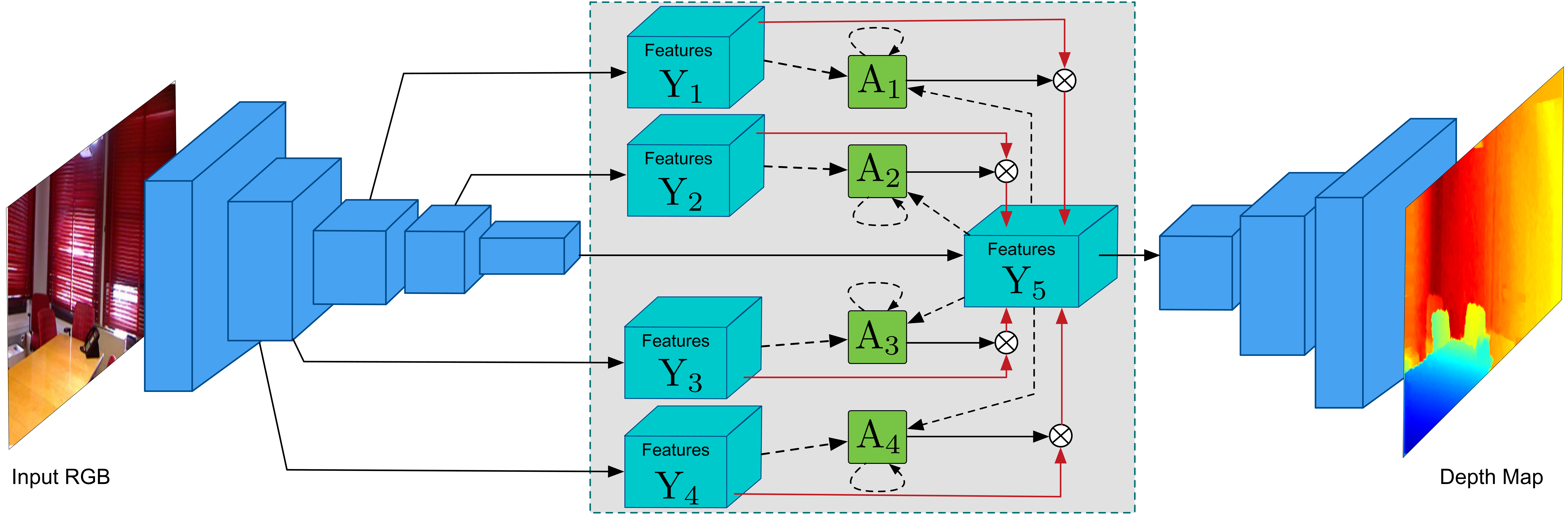} 
\caption{Illustration of the proposed network for monocular depth estimation. The blue blocks indicate the front-end CNN, which in our implementation is made
by an encoder and associated decoder (Section \ref{implementation}). The gray box contains a schematic representation of the proposed 
Structured Attention guided CRF model. Inside, the green boxes indicate the estimated attention maps, while the light blue ones represent the features jointly inferred with our CRF.
The arrows indicates the dependencies among the estimated variables used in our message passing algorithm (Section \ref{optimization}). The dashed arrows indicate the updates involving the attention model.}
\label{framework}
\vspace{-12pt}
\end{figure*}

In this section we describe our approach for estimating depth maps from still images.
We first provide an overview of our method and then introduce the proposed CRF model with structured attention. 
We conclude this section providing some details about our implementation. 

\subsection{Problem Formulation and Overview} 
As stated in the introduction, the problem of predicting a depth map from a single 
RGB image can be treated as a supervised learning problem. Denoting as $\mathcal{I}$ the space of RGB images and as 
$\mathcal{D}$ the domain of real-valued depth maps, given a training set 
$\mathcal{T} = \{(\vect{I}_i, \vect{D}_i)\}_{i=1}^M$, $\vect{I}_i \in \mathcal{I}$ and $\vect{D}_i \in \mathcal{D}$, we are interested in learning a non-linear mapping 
$\Phi:\mathcal{I} \rightarrow \mathcal{D}$. 

In analogy with previous works \cite{liu2015deep,xu2017multi}, we propose 
to learn the mapping $\Phi$ by building a deep architecture which is composed by two main building blocks: a front-end CNN 
and a CRF model. The main purpose of the proposed CRF model is to combine multi-scale information derived from the inner layers of the front-end CNN.
Differently from previous research \cite{liu2015deep,xu2017multi}, our CRF model does not simply act in order to refine the final prediction
map of the CNN neither requires as input multiple score maps of the same size. In this paper we argue that better estimates can be obtained
with a more flexible model which accepts as inputs a set of 
$S$ multi-scale feature maps $\vect{X} = \{\vect{X}_{s}\}_{s=1}^{S}$ derived directly from the front-end intermediate layers. 
To facilitate the modeling, all the multi-scale feature maps are resized to the same resolution via upsampling or downsampling operations. 
Here $\vect{X}_{s} = \{\vect{x}_{s}^i\}_{i=1}^{N}$, $\vect{x}_{s}^i \in \mathbb{R}^N$, indicates a set of feature vectors. 

The main idea behind the design of the proposed multi-scale CRF model is to estimate the depth map $\vect{D}$ associated to an RGB image
$\vect{I}$ by exploiting the features at the last layer $\vect{X}_{S}$ and a set of auxiliary feature representations derived
from the intermediate scales $s=1,\dots, S-1$. To do that, we propose to learn a set of latent feature maps 
$\vect{Y}_{s} = \{\vect{y}_{s}^i\}_{i=1}^{N}$, $s=1,\dots, S$ and to model the dependencies between the representations learned at the
last layer and those corresponding to each intermediate scales by introducing an appropriate attention model 
$\vect{A} = \{\vect{A}_{s}\}_{s=1}^{S-1}$, parameterized by binary variables
$\vect{A}_{s} = \{a_{s}^i\}_{i=1}^{N}$, 
$a_s^i \in \{0,1\}$. Intuitively, the attention variable $a_s^i$ regulates the information which is allowed to flow between
each intermediate scale $s$ and the final scale $S$ for pixel $i$. In other words, by learning the attention maps 
we automatically discover which information derived from inner CNN representations is relevant for final depth estimation. 
Furthermore, in order to obtain accurate attention maps $\vect{A}_{s}$ we propose to learn a structured attention model, \ie we impose structural
constraints on the estimated variables $a_s^i$ enforcing those corresponding to neighboring pixels to be related. 
Importantly, the proposed CRF jointly infers the hidden features and the attention maps.

Figure~\ref{framework} schematically depicts the proposed framework and our CRF model. The idea of modeling the relationships between
the learned representations at the finer scale and the features corresponding to each intermediate layer is inspired by the 
recent DenseNet architecture \cite{huang2016densely}. As demonstrated in our experiments (Section \ref{exp}), this strategy leads to improved performance with respect to
a cascade model as proposed in \cite{xu2017multi}.

\subsection{Structured Attention Guided Multi-Scale CRF}

\subsubsection{Proposed Model}
Given the observed multi-scale feature maps $\vect{X}$, we jointly estimate the latent 
multi-scale representations $\vect{Y}$ and the attention variables 
$\vect{A}$ by designing a Conditional Random 
Field model with the following associated energy function:
\begin{equation}
\setlength{\abovedisplayskip}{5pt}
 E (\vect{Y}, \vect{A}) = 
\Phi (\vect{Y}, \vect{X})   
+  \Xi (\vect{Y}, \vect{A}) + \Psi (\vect{A}) 
\label{equ:c-crf}
\setlength{\belowdisplayskip}{5pt}
\end{equation}
The first term in (\ref{equ:c-crf}) is the sum of unary potentials relating
the latent features representations $\vect{y}_{s}^i$ with the associated observations $\vect{x}_{s}^i$, \ie:
\begin{equation}
\setlength{\abovedisplayskip}{2pt}
 \Phi (\vect{Y}, \vect{X}) = \sum_{s=1}^S \sum_{i} \phi (\vect{y}_{s}^i, \vect{x}_{s}^i)  =  - \sum_{s=1}^S \sum_{i} \frac{1}{2}\|\vect{y}_{s}^i-\vect{x}_s^i\|^2
 \label{pot1}
\setlength{\belowdisplayskip}{2pt}
\end{equation}
As in previous works \cite{liu2015deep,xu2017multi} we consider Gaussian functions, such as to enforce the estimated latent features to be close
to their corresponding observations. 
The second term is defined as:
\begin{equation}
\setlength{\abovedisplayskip}{1pt}
 \Xi (\vect{Y}, \vect{A}) = \sum_{s\neq S} \sum_{i,j} \xi ({a}_{s}^{i}, \vect{y}_{s}^i, \vect{y}_{S}^j)
  \label{pot2}
\setlength{\belowdisplayskip}{1pt}
\end{equation}
It models the relationship between the latent features at the last scale with those
of each intermediate scale. This term also involves the attention variables ${a}_{s}^{i}$
which regulate the flow of information between related scales. We define: 
\begin{equation}
\setlength{\abovedisplayskip}{5pt}
 \xi({a}_{s}^{i},\vect{y}_{s}^i,\vect{y}_{S}^j) = {a}_{s}^{i}  \xi_y(\vect{y}_{s}^i,\vect{y}_{S}^j) = 
 {a}_{s}^{i} {\vect{y}}_{s}^i \vect{K}^{s}_{i,j} {\vect{y}}_{S}^j 
\setlength{\belowdisplayskip}{5pt}
\end{equation}
where $\vect{K}^{s}_{i,j}\in\mathbb{R}^{ C_s \times 
C_{S}}$ and $C_s$, $C_S$ refer to the number of channels of features scale $s$ and $S$, respectively. 
Finally, the third term in (\ref{equ:c-crf}) aims to enforce some structural constraints among attention variables. 
{For instance, it is reasonable to assume that the estimated attention maps
for related pixels and scales should be similar. To keep the computational cost limited, we only consider dependencies
among attention variables at the same scale and we define: }
\begin{equation}
\setlength{\abovedisplayskip}{5pt}
 \Psi (\vect{A}) = \sum_{s\neq S}\sum_{i,j} \psi ({a}_{s}^i, {a}_{s}^j) = \sum_{s\neq S}\sum_{i,j} \beta^{s}_{i,j}{a}_{s}^i a_{s}^j
\setlength{\belowdisplayskip}{5pt}
  \label{pot3}
\end{equation}
where $\beta^{s}_{i,j}$ are coefficients to be learned. To model dependencies between pairs of attention variables we consider
a bilinear function, in analogy with previous works \cite{kim2017structured}.

\vspace{-8pt}
\subsubsection{Deriving Mean-Field Updates}\label{learning}
Following previous works \cite{zheng2015conditional,xu2017multi} we resort on 
mean-field approximation. We derive mean-field inference equations for both latent features and attention variables. By denoting as
$\mathbb{E}_q$ the expectation with respect to the distribution $q$, we get:
\vspace{-9pt}
\begin{eqnarray}
 & q(\vect{y}_{s}^i) & \propto \exp\Big( \phi(\vect{y}_s^i,\vect{x}_s^i) + \\
 && \mathbb{E}_{q(a_{s}^i)}\{a_{s}^i\}  \sum_{j} 
 \mathbb{E}_{q(\vect{y}_{S}^j)}\{\xi_y(\vect{y}_{s}^i, \vect{y}_{S}^j)\} 
\Big), \nonumber \\
 & q(\vect{y}_{S}^i) &\propto \exp\Big( \phi(\vect{y}_S^i,\vect{x}_S^i) + \\ 
 &&\sum_{s\neq S}\sum_{j} 
\mathbb{E}_{q(a_{s}^j)}\{a_{s}^j\}  \mathbb{E}_{q(\vect{y}_{s}^j)}\{\xi_y(\vect{y}_{S}^i, \vect{y}_{s}^j)\} 
\Big), \nonumber \\
 &q(a_{s}^i) &\propto \exp\Big( a_{s}^i \mathbb{E}_{q(\vect{y}_{s}^i)}\Big\{ \sum_{j} 
\mathbb{E}_{q(\vect{y}_{S}^j)} \left\{\xi_y(\vect{y}_{s}^i, \vect{y}_{S}^j)\right\} \Big\} \label{a_updates}  \nonumber \\ 
&& + \sum_{s }\sum_{j} 
 \mathbb{E}_{q(a_{s}^j)}\{\psi (a_{s}^i, a_{s}^j)\} \Big), 
\end{eqnarray}
\vspace{-9pt}

By considering the potentials
defined in (\ref{pot1}), (\ref{pot2}) and (\ref{pot3}) and denoting as $\bar{a}_s^i=\mathbb{E}_{q(a_{s}^i)}\{a_{s}^i\}$ and
${\bar{\vect{y}}}_{s}^i=\mathbb{E}_{q({\vect{y}}_{s}^i)}\{{\vect{y}}_{s}^i\}$, the following mean-fields updates can be derived for the latent feature
representations:
\begin{equation}
\setlength{\abovedisplayskip}{2pt}
 \bar{\vect{y}}_{s}^i  =  \vect{x}_s^i + \bar{a}_s^i
\sum_{j} 
\vect{K}^{s}_{i,j} \bar{\vect{y}}_{S}^j 
\quad \label{MF1} 
\setlength{\belowdisplayskip}{2pt}
\end{equation}
\begin{equation}
\setlength{\abovedisplayskip}{2pt}
 \bar{\vect{y}}_{S}^i =  \vect{x}_S^i + \sum_{s\neq S} \sum_{j} \bar{a}_s^j
\vect{K}^{s}_{i,j} \bar{\vect{y}}_{s}^j 
\quad \label{MF2}
\setlength{\belowdisplayskip}{2pt}
\end{equation}
Since $a^i_s$ are binary variables, $\bar{a}_s^i = \frac{q(a_{s}^i=1)}{q(a_{s}^i=0)+q(a_{s}^i=1)}$. Therefore, 
the updates for $\bar{a}_s^i$ can be derived considering (\ref{a_updates}) and 
the definitions of potential functions (\ref{pot2}) and (\ref{pot3}):
\vspace{-2pt}
\begin{equation}
\setlength{\abovedisplayskip}{1pt}
 \bar{a}_{s}^i =  
\sigma \left( -\sum_{j} \bar{\vect{y}}_{s}^i
\vect{K}^{s}_{i,j} \bar{\vect{y}}_{S}^j - \sum_{s} \sum_{j} \beta^{s}_{i,j} \bar{a}_{s}^j \right) \quad \label{attention}
\setlength{\belowdisplayskip}{1pt}
\end{equation}
\vspace{-2pt}
where $\sigma()$ denotes the sigmoid function. Eqn. (\ref{attention}) shows that, in analogy with previous methods employing an attention 
model \cite{chen2015attention,hong2015TransferNet}, in our framework we also compute the attention variables
by applying a sigmoid function to the features derived by our CNN model. In addition, as we also consider dependencies among
different $a_s^i$ as in structured models \cite{kim2017structured}, our updates also involve related attention variables.

To infer the latent multi-scale representations $\vect{Y}$ and the attention variables 
$\vect{A}$, we implement the mean-field updates as a neural network (see Section \ref{optimization}). In this way
we are able to simultaneously learn the parameters of the CRFs and those of the front-end CNN. {When the inference is 
complete, the final depth map is obtained considering the final estimate associated to the last scale $\bar{\vect{y}}_{S}$ (see Section
\ref{implementation}).}

\begin{table*}[!t]
 \centering
 \caption{NYU Depth V2 dataset: comparison with state of the art. In bold we indicate the best method adopting on the original set, while in italics we
 indicate the best method using the extended set.}
\setlength\tabcolsep{8pt}
\resizebox{0.86\linewidth}{!} {
\begin{tabular}{l|c|ccc|ccc}
\toprule[1.3pt]
\multirow{2}{*}{Method} &\multirow{2}{*}{Extra Training Data ?} & \multicolumn{3}{c|}{\tabincell{c}{Error (lower is better)}} & \multicolumn{3}{c}{\tabincell{c}{Accuracy (higher is better)}} \\ \cline{3-8} & & rel & log10 & rms & $\delta < 1.25$ & $\delta < 1.25^2$ & $\delta < 1.25^3$ \\ \midrule\midrule
Saxena \etal~\cite{saxena2009make3d} &  No (795)   & 0.349 &  -     & 1.214  &0.447 & 0.745 & 0.897 \\
Karsch \etal~\cite{karsch2014depth}   &   No (795)  &0.35    &0.131&1.20   &    -     &     -    &    -      \\
Liu \etal~\cite{liu2014discrete} & No (795)   &  0.335 & 0.127 & 1.06  &    -     &     -    &     -     \\
Ladicky \etal~\cite{ladicky2014pulling}  &  No (795)   &   -       &    -    &    -     & 0.542& 0.829&  0.941 \\
Zhuo \etal~\cite{zhuo2015indoor}  &  No (795)    & 0.305 &0.122& 1.04  & 0.525& 0.838& 0.962 \\
Wang \etal~\cite{wang2015towards}     &    No (795)  & 0.220 & 0.094&0.745& 0.605 & 0.890 & 0.970 \\
Liu \etal~\cite{liu2016learningtpami}    &   No (795) & 0.213 &0.087& 0.759& 0.650& 0.906&  0.976 \\
Roi and Todorovic~\cite{roymonocular}         &      No (795)         & 0.187  &  0.078& 0.744& - &  -  &  -\\
Xu \etal \cite{xu2017multi} &        No (795)       & 0.139  & 0.063 & 0.609 &  0.793 & 0.948 & 0.984 \\ 
{Ours} &  No (795)   & \textbf{0.125}  & \textbf{0.057} & \textbf{0.593} & \textbf{0.806}  & \textbf{0.952} & \textbf{0.986} \\ \midrule\midrule
Eigen \etal~\cite{eigen2014depth} & Yes (120K)               & 0.215  &  -      & 0.907& 0.611 &  0.887  &  0.971\\
Eigen and Fergus~\cite{eigen2015predicting}    &   Yes (120K)       & 0.158  & -       & 0.641& 0.769 & 0.950 & 0.988 \\
Laina \etal~\cite{laina2016deeper}   & Yes (12K)             & 0.129  &0.056& {0.583}&0.801 & 0.950 & 0.986\\ 
Li \etal \cite{li2017monocular}  & Yes (24K) & 0.139 & 0.058 & 0.505 & 0.820   & 0.960 &   0.989 \\
Xu \etal \cite{xu2017multi} & Yes (12K)   & \textit{0.121}  & \textit{0.052}& \textit{0.586} & \textit{0.811}  & \textit{0.954} & \textit{0.987} \\
\bottomrule[1.3pt]                         
\end{tabular}
}
\label{sota_nyu}
\vspace{-10pt}
\end{table*}
\vspace{-10pt}
\subsubsection{Implementation with Neural Networks}
\label{optimization} 
To enable end-to-end optimization of the whole network, we implement the proposed multi-scale model in neural networks. The target is to perform mean-field 
updates for both the attention variables and the multi-scale feature maps
according to the derivation described in Section~\ref{learning}. 

To perform mean-field updates of the attention model $\vect{A}$ we follow (\ref{attention}). 
In practice, the update of each attention map $\vect{a}_s$ can be implemented in several steps as follows: (i) perform 
the message passing from the two associated feature maps $\vect{\bar{y}}_s$ and $\vect{\bar{y}}_S$ 
($\vect{\bar{y}}_s$ and $\vect{\bar{y}}_S$ are initialized with corresponding feature observations $\vect{x}_s$ and $\vect{x}_S$, respectively). 
The message passing is performed via convolutional operations as $\hat{\vect{a}}_s \leftarrow 
\vect{y}_{s} \odot (\vect{K}_{s} \otimes \bar{\vect{y}}_{S})$, where $\vect{K}_s$ is a convolutional kernel corresponding to the $s$-th scale and
the symbols $\otimes$ and $\odot$ denote the convolutional and the element-wise product 
operation, respectively; (ii) perform the message passing on the attention map with $\tilde{\vect{a}}_s \leftarrow \boldsymbol{\beta}_{s} \otimes \bar{\vect{a}}_s$, 
where $\boldsymbol{\beta}_s$ is a convolutional kernel; (iii) perform the normalization with sigmoid function$\bar{\vect{a}}_s \leftarrow 
\sigma(-(\hat{\vect{a}}_s \oplus \tilde{\vect{a}}_s))$, where $\oplus$ denotes element-wise addition operation. 

When the attention maps are updated, we use them as guidance to update the last scale feature map $\vect{y}_S$. The mean-field updates of $\vect{y}_S$ 
can be carried out according to (\ref{MF2}) as follows: (i) perform the message passing from the $s$-th scale to the $S$-th scale by $\hat{\vect{y}}_{s}
\leftarrow 
\vect{K}_{s} \otimes \bar{\vect{y}}_{s}$; (ii) multiply for the attention model and add the unary term $\mathbf{x}_{S}$ by 
$\bar{\vect{y}}_{S} \leftarrow \mathbf{x}_{S} \oplus  \sum_{s} (\bar{\vect{a}}_s \odot \hat{\vect{y}}_{s})$.  
The computation of mean-field updates for the latent features corresponding to intermediate scales can be performed similarly, according to (\ref{MF1}). 
In our implementation to reduce the computational overhead, we do not perform the mean-field updates for the intermediate scales. 
The attention maps and the last scale feature map are iteratively updated.  

We would like to remark that, as a consequence of the definition of the potential functions in (\ref{pot1}), (\ref{pot2}) and (\ref{pot3}),
the computations of the mean-field updates in our approach are much more efficient than in \cite{xu2017multi} where Gaussian functions
are considered for pairwise potentials. 
Indeed, Gaussian convolutions involve a much higher computational overhead both in the forward and in the backward pass. 
We further discuss this aspect in Section \ref{exp}. 
\vspace{-3pt}
\subsection{Network Structure and Optimization}
\vspace{-3pt}
\label{implementation}
\paragraph{Network Structure and Implementation.} The overall framework for monocular depth estimation is made by a CNN architecture and the proposed CRF model (Fig.~\ref{framework}).
The CNN architecture is made of two main components, \ie a fully convolutional encoder and a fully convolutional decoder. 
The encoder naturally supports any network structure. In this work we specifically employ ResNet-50~\cite{he2016deep}. In our implementation the proposed 
CRF is adopted to refine the last scale feature map derived from the semantic layer \textit{res5c}, which receives 
message from the other scale feature maps derived from \textit{res3c} and \textit{res4f}. $res3c$, $res4f$ and $res5c$ are the last layers of different convolutional 
blocks. In each convolutional block, every layer outputs a feature map with the same number of channels. Before message passing, all the feature maps are 
first upsampled using a deconvolutional operation to the same size, \ie~1/4 resolution of the original input image, and the number of channels is set to 256 
for all of them. The kernel size for both $\vect{K}_s$ and $\boldsymbol{\beta}_{s}$ is set to 3 with stride 1 and padding 1 to have a local receptive field and to
speed up the calculation.

The proposed multi-scale CRF module outputs a refined feature map.
To obtain the final prediction we upsample the feature map to the original resolution as the input image using deconvolutional operations. 
Each time we upsample the feature map by a factor of 2, at the same time reducing by half the number of feature channels.
\vspace{-10pt}
\paragraph{End-to-end optimization.} As stated above, the proposed model can be trained end-to-end, \ie the parameters of the front-end encoder $\vect{\Theta}_e$, those 
associated to the structured attention guided CRF $\vect{\Theta}_c$, and those of the decoder $\vect{\Theta}_d$ can be jointly optimized. 
Given the training data set ${\cal T}$, following previous works \cite{xu2017multi}, we use a square loss function for the optimization, \ie:
\vspace{-2pt}
\begin{equation*}
\setlength{\abovedisplayskip}{2pt}
\mathcal{L}_F((\mathcal{I}, \mathcal{D}; \vect{\Theta}_e,\vect{\Theta}_c, \vect{\Theta}_d)=\sum_{i=1}^{M}\|F(\vect{I}_i^l; 
\vect{\Theta}_e,\vect{\Theta}_c, \vect{\Theta}_d) - \vect{D}_i^l\|_2^2 \nonumber
\setlength{\belowdisplayskip}{2pt}
\end{equation*}
\vspace{-2pt}
The whole network is jointly optimized via back-propagation with standard stochastic gradient descent.
\vspace{-5pt}
\section{Experiments}
\vspace{-5pt}
\label{exp}
\begin{figure*}[t]
\centering
\includegraphics[width=.93\linewidth]{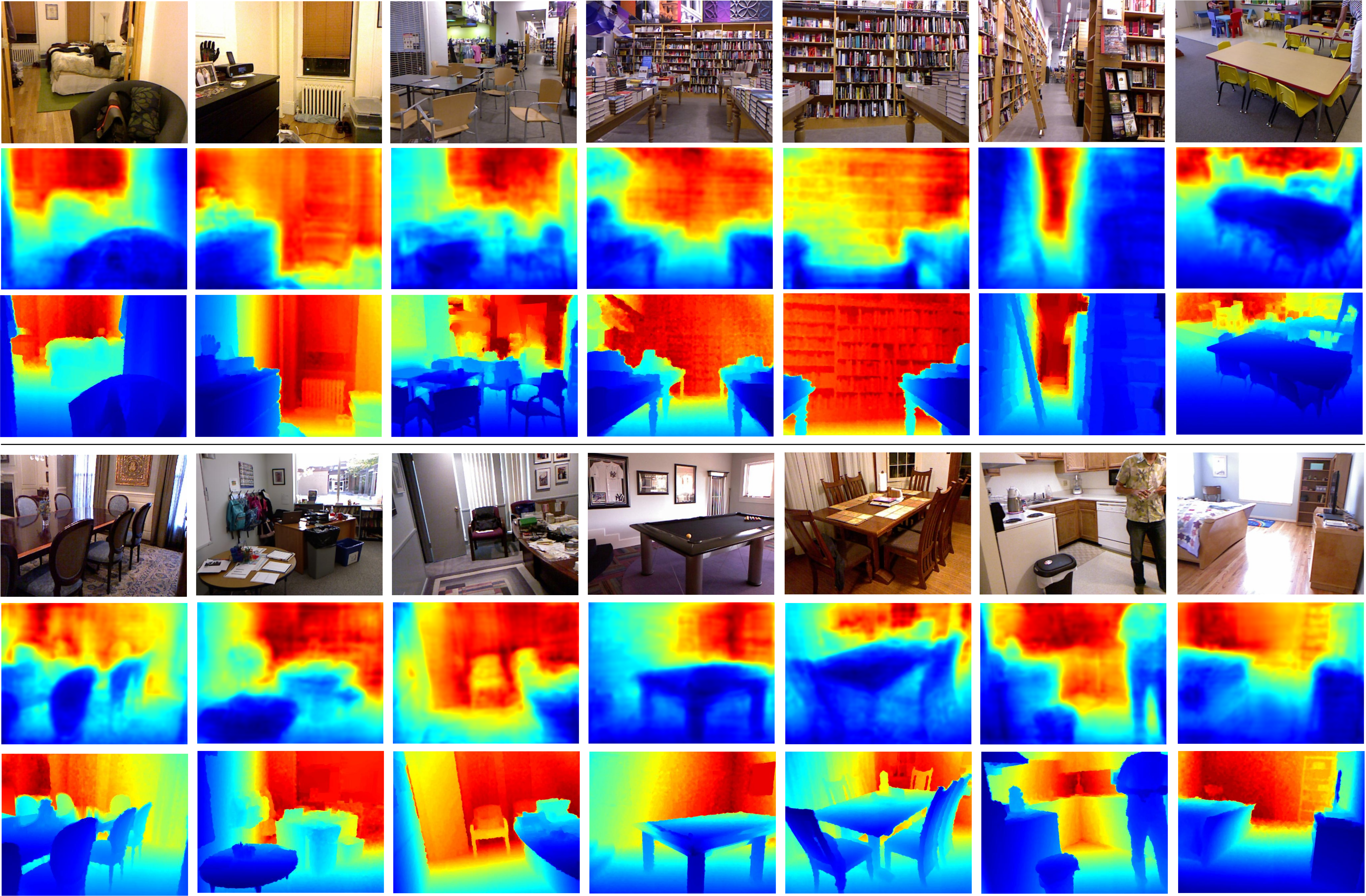} 
\caption{Examples of predicted depth maps on the NYU V2 test dataset: original RGB images (top row), 
predicted depth maps (center) and ground truth (bottom row).}
\label{nyud_examples}
\vspace{-6pt}
\end{figure*}

\begin{table*}
\centering
\caption{KITTI dataset: comparison with state of the art. In bold we indicate the best performances in the monocular setting, while in italics 
those corresponding to the stereo setting.}
\setlength\tabcolsep{8pt}
\resizebox{0.86\linewidth}{!} {
\begin{tabular}{l|cc|ccc|ccc}
\toprule[1.3pt]  
\multirow{2}{*}{Method} & \multicolumn{2}{c|}{Setting} 
& \multicolumn{3}{c|}{\tabincell{c}{Error (lower is better)}} & \multicolumn{3}{c}{\tabincell{c}{Accuracy (higher is better)}} \\\cline{2-9}
                                      & range & stereo? 
                                      & rel & sq rel & rms & $\delta < 1.25$ & $\delta < 1.25^2$ & $\delta < 1.25^3$ \\\midrule\midrule
Saxena \etal~\cite{saxena2009make3d}   & 0-80m & No 
& 0.280  &  - & 8.734 & 0.601  &  0.820  &  0.926\\
Eigen \etal~\cite{eigen2014depth}    & 0-80m & No 
& 0.190  &  -      &  7.156 &  0.692 &  0.899  &  0.967 \\
Liu \etal~\cite{liu2016learningtpami}    & 0-80m & No 
& 0.217  &  0.092   &  7.046 &  0.656 &  0.881  &  0.958 \\
Zhou~\etal~\cite{zhou2017unsupervised} & 0-80m & No  
& 0.208   &  1.768  & 6.858  & 0.678   & 0.885   & 0.957  \\
Kuznietsov \etal \cite{kuznietsov2017semi} (only supervised) & 0-80m & No & - & - & 4.815 & \textbf{0.845} & \textbf{0.957} & \textbf{0.987} \\
Ours  & 0-80m & No 
&\textbf{ 0.122} & \textbf{0.897} & \textbf{4.677}  & {0.818} & {0.954} & {0.985} \\ \midrule\midrule
Garg~\etal~\cite{garg2016unsupervised}  & 0-80m & Yes 
& 0.177 & 1.169  & 5.285    &  0.727 & 0.896  &  0.962   \\
Garg~\etal~\cite{garg2016unsupervised} L12 + Aug 8x  & 1-50m & Yes 
& 0.169  & 1.080   & 5.104    &  0.740 & 0.904  &  0.958   \\
Godard \etal~\cite{godard2016unsupervised}   & 0-80m & Yes 
& 0.148 & 1.344  & 5.927    &  0.803 & 0.922  &  0.963   \\
Kuznietsov \etal \cite{kuznietsov2017semi} & 0-80m & Yes & - & - & \textit{4.621} & \textit{0.852} & \textit{0.960} & \textit{0.986} \\ 
\bottomrule[1.3pt]                             
\end{tabular}
}
\label{sota_KITTI}
\vspace{-15pt}
\end{table*}

We demonstrate the effectiveness of the proposed approach performing experiments on two publicly available datasets: 
the NYU Depth V2 ~\cite{silberman2012indoor} and the KITTI~\cite{Geiger2013IJRR} datasets. The following subsections describe
our experimental setup and the results of our evaluation.  

\subsection{Experimental Setup}
\label{setup}
\paragraph{Datasets.}
The {NYU Depth V2} dataset~\cite{silberman2012indoor} has 120K pairs of RGB and depth maps gathered with a 
Microsoft Kinect. The image resolution is
$640 \times 480$ pixels. 
The dataset is split into a training (249 scenes) and a test set (215 scenes). Following previous works~\cite{liu2015deep,zhuo2015indoor,xu2017multi} in our
experiments we consider a subset of 1449 RGB-D pairs, of which 795 are used for training and the rest for testing. The data augmentation is performed on the 
fly by cropping the images to $320 \times 240$ pixels, randomly flipping and scaling them with a ratio $\rho \in \{1, 1.2, 1.5\}$.

The {KITTI} dataset~\cite{Geiger2013IJRR}, originally built for testing computer vision algorithms in several tasks in the context of autonomous driving, 
contains depth images captured with a LiDAR sensor mounted on a driving vehicle. In our experiments we follow the experimental protocol 
proposed by Eigen~\etal~\cite{eigen2014depth} and consider 22,600 images corresponding to 32 scenes as training data
and 697 images asoociated to other 29 scenes as test data. The RGB image resolution is reduced by half with respect to the original $1224\times 368$ pixels.
The ground-truth depth maps are generated by reprojecting the 3D points collected from velodyne 
laser into the left monocular camera as detailed in~\cite{garg2016unsupervised}. 
\vspace{-15pt}
\paragraph{Implementation Details.}
The proposed approach is implemented using the Caffe framework \cite{jia2014caffe} 
and runs on a single Nvidia Titan X GPU with 12 GB memory. While the proposed
framework is general, following recent works \cite{xu2017multi,laina2016deeper},
we adopt the ResNet50~\cite{he2016deep} as the front-end network architecture. As stated above in the implementation, 
we consider three-level feature maps derived from different semantic convolutional layers (\ie $res3c$, $res4f$ and $res5c$). 
These feature maps are fused with the proposed CRF model for the final prediction of the depth map. 
During training, the front-end network is initialized with ImageNet pretrained parameters. Differently from~\cite{xu2017multi} 
which requires a pretraining phase of the front-end CNN, we jointly optimize the whole network. The initial 
learning rate is set to $10e-9$, and is decreased 10 times every 40 epochs. In total 60 epochs are used for training. 
The mini-batch size is set to 16. The weight decay and the momentum are 0.0005 and 0.99, respectively.   

%


\begin{figure}[t]
\centering
\includegraphics[width=.99\linewidth]{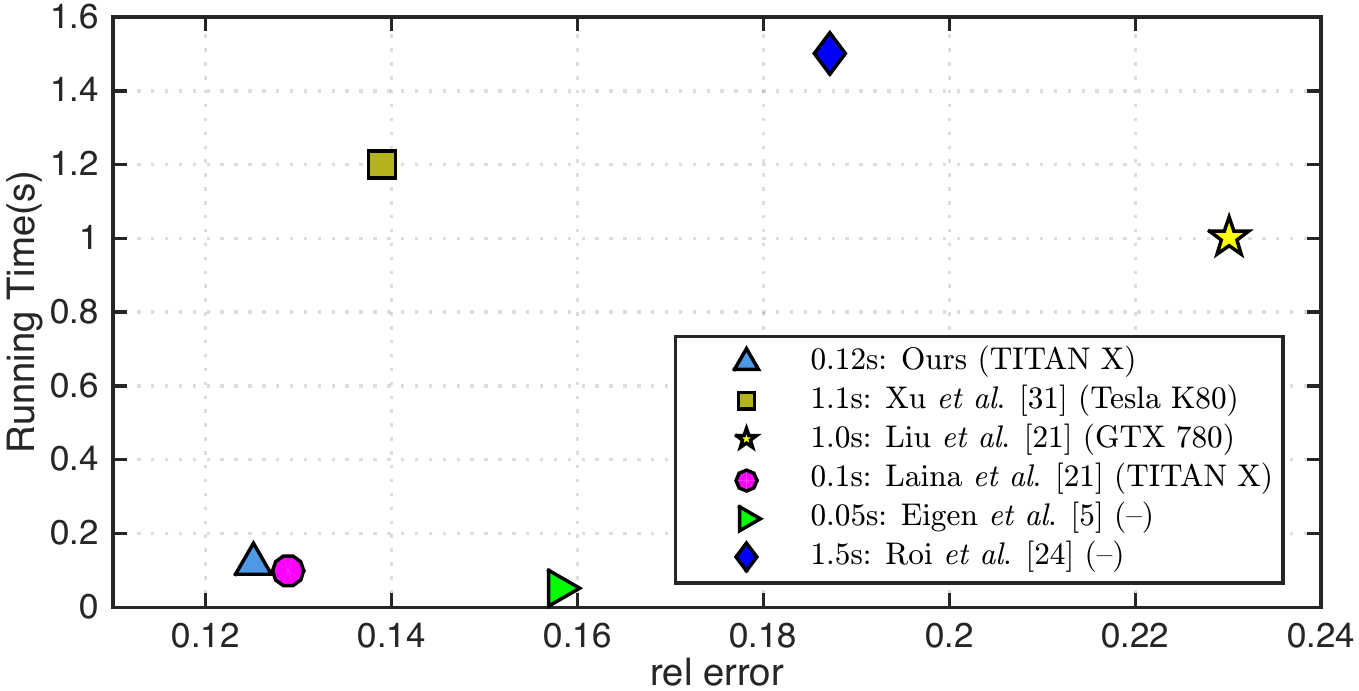} 
\caption{NYU V2 dataset. Comparison with previous methods: running time vs. rel error.}
\label{time_comp}
\vspace{-5mm}
\end{figure}


\begin{figure*}[t]
\centering
\includegraphics[width=.99\linewidth]{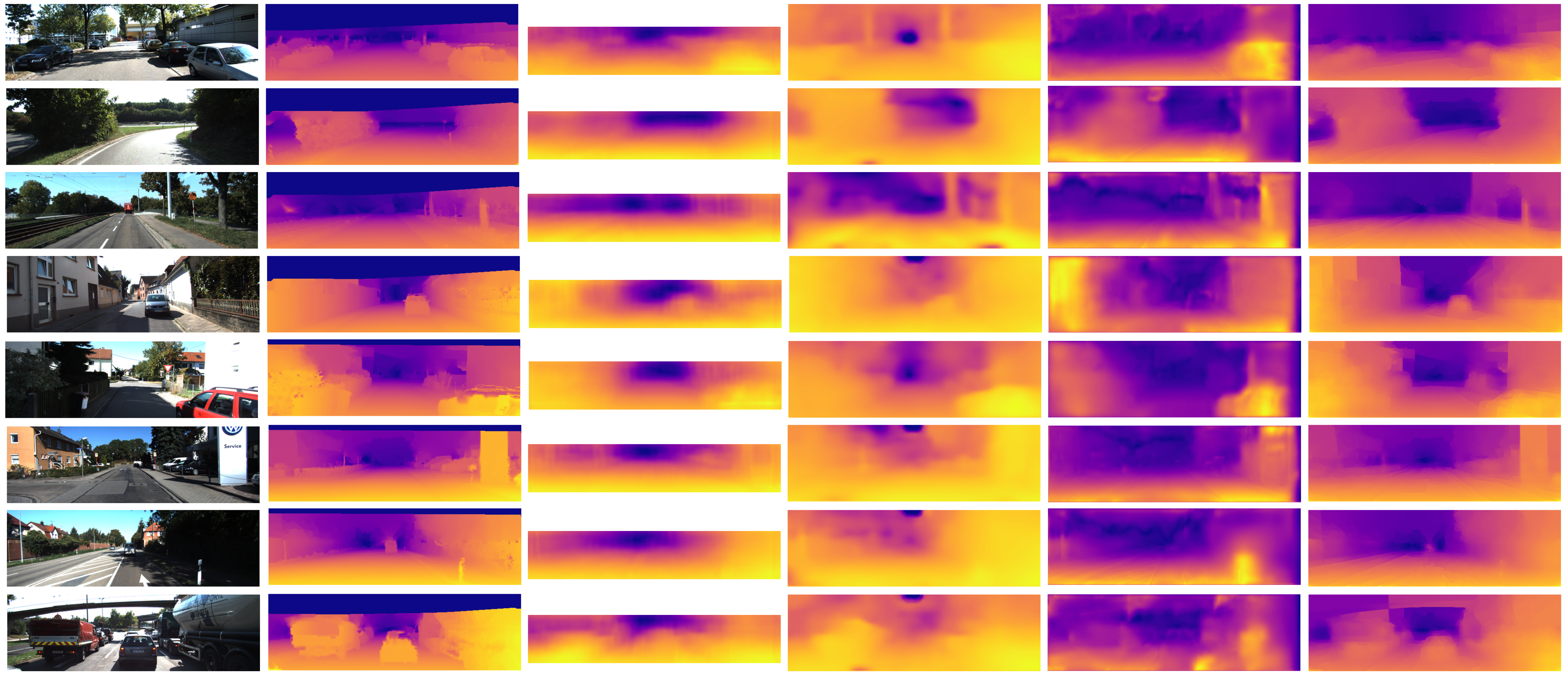} 
\put(-476,215){\footnotesize RGB Image}
\put(-395,215){\footnotesize GT Depth Map}
\put(-310,215){\footnotesize Eigen~\etal~\cite{eigen2014depth}}
\put(-230,215){\footnotesize Zhou~\etal~\cite{zhou2017unsupervised}}
\put(-145,215){\footnotesize Garg~\etal~\cite{garg2016unsupervised}}
\put(-51,215){\footnotesize Ours}
\caption{Examples of depth prediction results on the KITTI dataset: qualitative comparison with previous methods. 
The sparse ground-truth depth maps are interpolated for better visualization.}
\label{KITTI_results}
\vspace{-5pt}
\end{figure*}
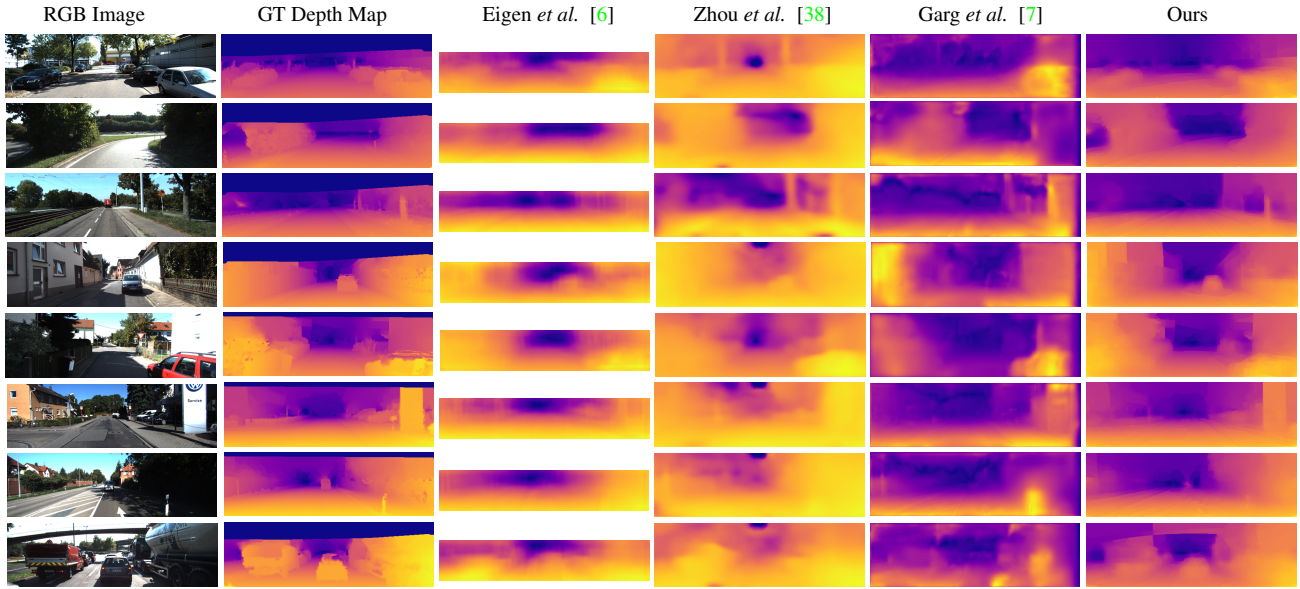
\par\textbf{Evaluation Metrics.} In analogy with previous works~\cite{eigen2015predicting,eigen2014depth,wang2015towards,xu2017multi}, 
to quantitatively assess the performance of our method we consider several evaluation metrics.
Specifically if $Q$ is the total number of pixels of the test set and $\bar{d}_i$ and $d_i$ denote the estimated 
and the ground-truth depth for pixel $i$, we compute:
(i) the mean relative error (rel): 
\( \frac{1}{Q}\sum_{i=1}^Q\frac{|\bar{d}_i - d_i|}{d_i} \); (ii) the
root mean squared error (rms): 
\( \sqrt{\frac{1}{Q}\sum_{i=1}^Q(\bar{d}_i - d_i)^2} \); (iii) the
mean log10 error (log10): 
\( \frac{1}{Q}\sum_{i=1}^Q \Vert \log_{10}(\bar{d}_i) - \log_{10}(d_i) \Vert \) and (iv) the 
accuracy with threshold $t$, \ie the percentage of $\bar{d}_i$ such that $\delta=\max (\frac{d_i}{\bar{d}_i}, \frac{\bar{d}_i}{d_i}) < t$, 
where $t \in [1.25, 1.25^2, 1.25^3]$.

In order to compare our results with previous methods on the KITTI dataset 
we crop our images using the evaluation crop applied by Eigen \etal \cite{eigen2014depth}.

\begin{table*}
\centering
\caption{Quantitative analysis of the main components of our method on the KITTI dataset. 
}

\Huge
\setlength\tabcolsep{10pt}
\resizebox{0.88\linewidth}{!} {
\begin{tabular}{l|ccc|ccc}
\toprule[2.5pt]
\multirow{2}{*}{\tabincell{c}{Method}} & \multicolumn{3}{c|}{\tabincell{c}{Error \\ (lower is better)}} & \multicolumn{3}{c}{\tabincell{c}{Accuracy \\ (higher is better)}} \\\cline{2-7}
                                      & rel & log10 & rms & $\delta < 1.25$ & $\delta < 1.25^2$ & $\delta < 1.25^3$ \\\midrule\midrule
Front-end CNN (w/o multiple deep supervision)            & 0.168& 1.072& 5.101& 0.741 & 0.932 & 0.981\\ 
Front-end CNN (w/ multiple deep supervision)            &  0.152 & 0.973 &  4.902 & 0.782  &  0.931 & 0.974 \\
Multi-scale feature fusion with naive concatenation   & 0.143  & 0.949 &  4.825 & 0.795 & 0.939 & 0.978  \\
Multi-scale feature fusion with CRFs (w/o attention model)  & 0.134  & 0.895 & 4.733 & 0.803  & 0.942  & 0.980 \\
Multi-scale feature fusion with CRFs (w/ attention model)   & 0.127  & 0.869 & 4.636 & 0.811 & 0.950 & 0.982 \\
Multi-scale feature fusion with CRFs (w/ structured attention model)  &  0.122 & 0.897 & 4.677  & 0.818 & 0.954 & 0.985  \\
\bottomrule[2.5pt]                            
\end{tabular}
}
\label{cnn_arch}
\vspace{-15pt}
\end{table*}


%
%
\subsection{Experimental Results}
To demonstrate the effectiveness of the proposed framework we first conduct a 
comparison with state of the art methods both on the NYU Depth V2 dataset and on the KITTI benchmark.
We also conduct an in-depth analysis of our method, evaluating both accuracy and
computational efficiency.
\vspace{-3mm}

\paragraph{NYU Depth V2 Dataset.} Table~\ref{sota_nyu} shows the results of the comparison with state of the art methods on the NYU Depth V2 dataset.
As baselines we consider both approaches based on hand-crafted features (Saxena \etal~\cite{saxena2009make3d}, Karsch \etal~\cite{karsch2014depth},
Ladicky \etal~\cite{ladicky2014pulling}) and deep learning
architectures. Concerning the latter category, we compare with 
methods which exploit multiscale information (Eigen \etal~\cite{eigen2014depth}, Eigen and Fergus~\cite{eigen2015predicting}, Li \etal \cite{li2017monocular}), with 
approaches which consider graphical models (Liu \etal~\cite{liu2014discrete}, Liu \etal~\cite{liu2015deep}, Zhuo \etal~\cite{zhuo2015indoor}, 
Wang \etal~\cite{wang2015towards}, Xu \etal \cite{xu2017multi}) and neural regression forests (\cite{roymonocular}),
and with methods which explore the utilization of the reverse Huber loss function (Laina \etal~\cite{laina2016deeper}). The numerical results associated to previous methods are taken directly from the original papers. For a fair comparison
in the table we also report information about the adopted training set, 
as it represents an important factor for CNN performance. In particular, we separate methods 
which adopt the original training set in \cite{eigen2014depth} and those which consider an extended dataset for
learning their deep models.

Our results clearly show that the proposed approach outperforms all supervised learning
methods adopting the original dataset in \cite{eigen2014depth}.
Importantly, the performance improvements over previous works based on CRFs models \cite{wang2015towards,liu2015deep,xu2017multi}
are significant. In particular, we believe that the increase in accuracy with respect to \cite{xu2017multi} confirms our initial intuition that
operating directly at feature-level and integrating an attention model into a CRF leads to more accurate depth estimates. 
Finally, we would like to point out that our approach also outperforms most methods considering an extended
training set. Furthermore, the performance gap between our framework and the deep model in \cite{xu2017multi} trained on 95K samples  
is very narrow. We also provide some examples of depth maps estimated with the proposed
method in Fig.~\ref{nyud_examples}. Comparing our prediction with ground truth it is clear that our approach 
is quite accurate even at objects boundaries (notice, for instance, the accuracy in recovering fine-grained details in case of objects like chairs and tables). 

Finally, we compare the proposed approach with previous methods considering the computational cost in the test phase. 
Figure \ref{time_comp} depicts the mean relative error vs the running time (\ie time to classify one image) for some of the baseline methods (numbers
are taken from the original papers). Our approach guarantees the best trade-off between accuracy and time (notice that the deep model in Laina 
\etal \cite{laina2016deeper} is trained on an extended dataset).
It is interesting to compare our method with \cite{xu2017multi}: the proposed framework not only outperforms \cite{xu2017multi} in terms of accuracy when both models are
trained on the original set \cite{eigen2014depth}) but, by adopting different potential functions in the CRF, results into a much faster
inference. Another interesting comparison is with \cite{wang2015towards} and \cite{liu2015deep}, as these works are also based on CRFs.
Our model significantly outperforms \cite{liu2015deep} and \cite{wang2015towards} both in terms of accuracy and of running time (see Fig.\ref{time_comp} and Table \ref{sota_nyu}):
due to visualization issues we do not show \cite{wang2015towards} in Fig.\ref{time_comp} as the original paper report
a time of 40 seconds to recover the depth map for a single image. 

\vspace{-3mm}

\paragraph{KITTI Dataset.} A comparison with state of the art methods is also conducted on the KITTI dataset and the associated results are shown in Table~\ref{sota_KITTI}. 
As baselines we consider the work by Saxena \etal~\cite{saxena2005learning}, 
Eigen \etal~\cite{eigen2014depth}, Liu \etal~\cite{liu2015deep}, Zhou~\etal~\cite{zhou2017unsupervised}, Garg~\etal~\cite{garg2016unsupervised}, 
Godard~\etal~\cite{godard2016unsupervised} and Kuznietsov \etal \cite{kuznietsov2017semi}. 
Importantly, the first four methods only employ monocular images to predict depth information, while in
\cite{garg2016unsupervised}, \cite{godard2016unsupervised} and \cite{kuznietsov2017semi} 
a stereo setting is considered in training and therefore these methods are not directly comparable with
our approach. As shown in the table, our approach outperforms all previous methods considering a supervised setting
with the exception of the recent method in \cite{kuznietsov2017semi}. With respect to \cite{kuznietsov2017semi} we obtain a
lower error, while the accuracy is slightly inferior. For sake of completeness we also report the performance of previous methods
considering a stereo setting. Among these methods, Kuznietsov \etal \cite{kuznietsov2017semi} achieve the best performance by exploiting
both ground truth supervision and stereo information. Following the same idea, we believe that an interesting future research direction will be to
integrate stereo cues into our framework.
A qualitative comparison with some state of the art methods is also shown in Fig.~\ref{KITTI_results}.

%

\vspace{-15pt}
\paragraph{Ablation Study.} To further demonstrate the effectiveness of the proposed method we conduct an ablation study on the KITTI dataset. Table \ref{cnn_arch} shows the
results of our analysis. In the table, ``multiple deep supervision" refers to training the front-end CNN with the approach in~\cite{xie2015holistically}; ``w/ attention model" refers to considering attention variables $a^i_s$ in the optimization but discarding the structured potential; ``w/ structured attention model'' indicates the using of the structured attention model. In line with findings from previous works \cite{wang2015towards,xu2017multi,liu2015deep}, embedding a CRFs model into
a deep architecture provides a significant improvement in terms of performance. Furthermore, adopting a CRFs is an extremely effective strategy for 
combining multi-scale features, as it is evident when comparing our results with CRF and those corresponding to naive feature concatenation. Finally and more importantly,
by introducing the proposed CRF model with an attention mechanism and, in particular, with a structured attention one, we can significantly boost performance.

\vspace{-5pt}
\section{Conclusions}
\vspace{-5pt}
We presented a novel approach for monocular depth estimation. The main contribution of this work
is a CRF model which optimally combines multi-scale information derived from the inner layers of a 
CNN by learning a set of latent features representations and the associated attention model.
We demonstrated that by combining multi-scale information at feature-level and by
adopting a structured attention mechanism, our approach significantly outperforms previous depth estimation
methods based on CRF-CNN models \cite{zheng2015conditional,liu2015deep,xu2017multi}.
Importantly, our framework can be used in combination with several
CNN architectures and can be trained end-to-end. Extensive evaluation shows that our method outperforms most baselines. Future research could perform cross-domain detection tasks~\cite{xu2017learningcross} based on the prediction of the scene depth.

\vspace{-5pt}
\section*{Acknowledgements}
\vspace{-5pt}
This work is partially supported by National Natural Science Foundation of China (NSFC, No.U1613209). The authors would like to thank NVIDIA for GPU donation. 

{\small
\bibliographystyle{ieee}
\bibliography{egbib}
}

\end{document}